# Security for Machine Learning-based Systems: Attacks and Challenges during Training and Inference

Faiq Khalid[*], Muhammad Abdullah Hanif [*], Semeen Rehman[†], Muhammad Shafique[*]
[*]*Institute of Computer Engineering, Vienna University of Technology (TU Wien), Austria*
[†]*Institute of Computer Technology, Vienna University of Technology (TU Wien), Austria*
{faiq.khalid; muhammad.hanif; semeen.rehman; muhammad.shafique}@tuwien.ac.at

***Abstract* — The exponential increase in dependencies between the cyber and physical world leads to an enormous amount of data which must be efficiently processed and stored. Therefore, computing paradigms are evolving towards machine learning (ML)-based systems because of their ability to efficiently and accurately process the enormous amount of data. Although ML-based solutions address the efficient computing requirements of big data, they introduce (new) security vulnerabilities into the systems, which cannot be addressed by traditional monitoring-based security measures. Therefore, this paper first presents a brief overview of various security threats in machine learning, their respective threat models and associated research challenges to develop robust security measures. To illustrate the security vulnerabilities of ML during training, inferencing and hardware implementation, we demonstrate some key security threats on ML using LeNet and VGGNet for MNIST and German Traffic Sign Recognition Benchmarks (GTSRB), respectively. Moreover, based on the security analysis of ML-training, we also propose an attack that has a very less impact on the inference accuracy. Towards the end, we highlight the associated research challenges in developing security measures and provide a brief overview of the techniques used to mitigate such security threats.**

## I. Introduction

Due to the exponential growth in complex integration and interconnection of the physical and cyber domains with humans, the number of connected devices increases enormously. Several market analysts and corresponding surveys forecast that by 2025, the connected devices (20 billion in 2017) will surpass 75 billion [1][2][3], as shown in Fig. 1. Although these connected devices are revolutionizing several applications domains like healthcare, industrial automation, autonomous vehicles, transport systems and many other but they generate and collect an enormous amount of data, for example, on average, 300 hours of video content is uploaded on YouTube every minute [4], 95 million [5] and 300 million [5] photos are uploaded daily on Instagram [4] and Facebook [4], respectively, as shown in Fig. 1. Moreover, several surveys also predict that by 2025, this data is expected to surpass the 160 Zettabyte mark and by 2020, for every person on earth, 1.7 MB data will be generated every second [6]. Therefore, there is a dire need to efficiently process and store this enormous amount of data, which leads to the following fundamental research challenges:

1) How to *increase the computing capability* to process this data *in a highly energy-efficient manner?*
2) How to *increase the storing capability* to store this data in interpretable form *with minimum energy and area overhead?*

To address these research challenges, researchers have been exploring several novel computing architectures, methodologies, frameworks, algorithms and tools. However, Machine learning (ML) algorithms, especially (deep) neural networks have emerged as one of the most popular computing paradigms due to their ability to efficiently handle such gigantic amounts of data [3]. ML algorithms not only addressed the processing requirements of the huge data but they have also revolutionized several application domains, like smart cities, intelligent transportation systems, smart grids, autonomous vehicles, healthcare, social networks (Facebook, Twitter, Youtube, Instagram, etc.) and many more, as shown in Fig. 2, by extracting some of the hidden features.

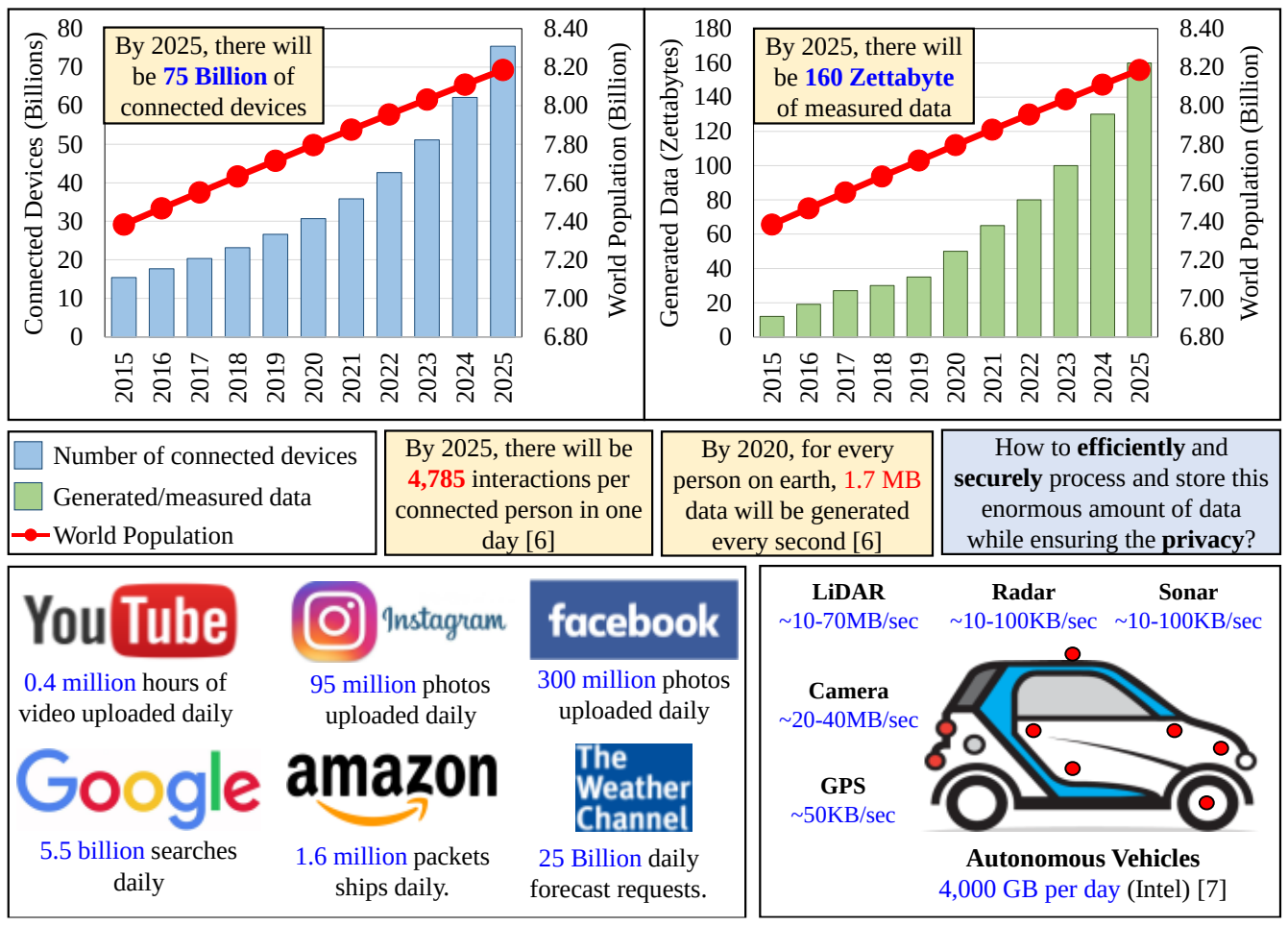

**Fig. 1: Increasing trend of connected devices and corresponding measured data for processing [2] (Source for logos: google images).**

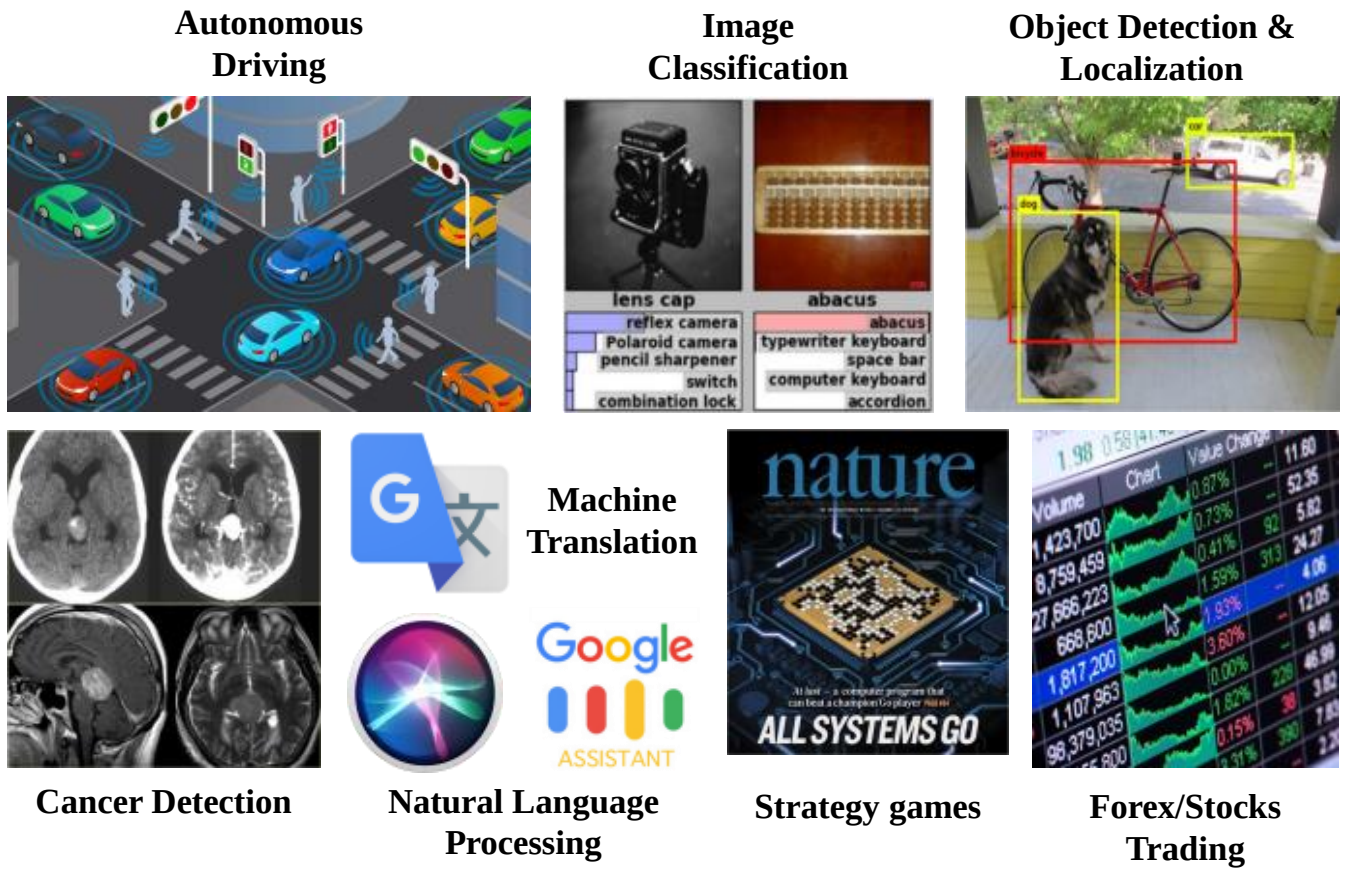

**Fig. 2: Applications of Machine Learning Algorithms (Source for images: google images).**

Unlike the traditional computing algorithms, ML algorithms dynamically change the computational flow with respect to the input data, which increases the energy overhead. Moreover, due to the unpredictability of the computing in hidden layers of neural networks, these algorithms possess several security vulnerabilities which result in increased system vulnerability towards security threats [8]-[11]. Some example are: Amazon echo hacking [8], Facebook chatbots [8], self-driving bus crashes (on its very first day in Las Vegas) [13]. These real-world incidents highlight the security risks involved in ML-based systems and raise the fundamental research question: *How to securely train and implement ML algorithms?*

To address this question, a comprehensive set of vulnerability analyses and countermeasures is required. Several techniques for security vulnerability analysis and countermeasures have been proposed based on the manufacturing/design cycle of ML-based systems which is composed of *training*, *hardware implementation* of trained model (e.g., LeNet and VGGNet trained for MNIST and GTSRB dataset) and *inference*. Since, each stage possesses its own security vulnerabilities, their corresponding methodologies for security analysis and countermeasures have different impact on ML-based systems. For example, traffic sign detection in autonomous vehicles requires a sophisticated and powerful pre-processing methodology to remove random environmental uncertainties [14]. However, face detection inside a highly secure building requires a relatively less powerful preprocessing methodology because of the controlled environment.

### A. *Our Contributions*

To address these fundamental security challenges in ML, this paper makes the following contributions:

1) A brief yet *comprehensive overview of the machine learning security* including various security threats at different stages (Section II.B) and respective threat models (Section II.A).
2) *Associated research challenges to develop robust security measures* for security threats at different stages of machine learning.
3) A comprehensive experimental analysis of different *security vulnerabilities* and *potential threats* during training and inference.
4) A brief discussion on the *possible countermeasures* for security vulnerabilities in machine learning.

## II. Security for Machine Learning

To provide a better understanding of the security for ML-based systems, in this section, we provide a comprehensive overview of possible threat models and associated security attacks/vulnerabilities during the different stages of the design/manufacturing cycle of the ML-based systems.

### A. *Threat Models*

To develop security measures for ML-based systems, the foremost step is to identify the potential threat factors, i.e., attacker, design/manufacturing stage, attack mechanism and intention of attack. Therefore, a precise threat model should be defined, which provides the information about the capabilities and goals of an attacker under realistic assumptions. Hence, first we provide a brief overview of the manufacturing cycle of ML-based applications/systems.

The manufacturing/design cycle is defined as all the possible steps which are involved in training, testing and deployment of the ML-based application/systems, as shown in Fig. 3 [15][17]. Based on the different resource requirements and potential application users, the following actors are part of the manufacturing/design cycle.

1) **3rd Party (3P) Cloud Platforms:** If the IP providers or manufacturers do not have enough computing resources to fulfill the requirements of the larger datasets and neural networks, e.g., ResNet [16], then 3rd party (3P) cloud platforms are used [17]. However, it comes with security vulnerabilities, i.e., IP stealing, manipulation of the training dataset and models/architectures. Therefore, cloud platforms can be declared as untrusted in the following two different cases.
   a) If the cloud platform provider is untrusted, it can manipulate the training dataset and baseline neural networks or ML algorithms.
   b) Even if the cloud platform provider is trusted, a man-in-middle [18] attack can be performed by another client to steal the IP, i.e., the trained network or even to manipulate the IP or affect the training process.
2) **IP Providers:** The other actor in the manufacturing cycle is the IP provider which can also be untrusted because it can poison the training datasets and can also manipulate baseline ML models/architectures or other hyper-parameters, which are not accessible to 3P cloud providers.
3) **Manufacturers:** If the manufacturers are untrusted then the following security vulnerabilities can be introduced:
   a) Malicious hardware during the hardware implementation.
   b) Side-channel (SC) attacks to steal the IP which can be a trained ML algorithm or a dataset.
   c) Trained algorithm by performing local training.
4) **Users:** Even after the deployment of the ML-based systems, the following security vulnerabilities can still be exploited:
   a) Users can perform side-channel attacks to extract the IP, i.e., trained ML algorithm.
   b) During inference, an attacker can also compromise the security of the ML-based system by manipulating the inference data or their corresponding hardware.

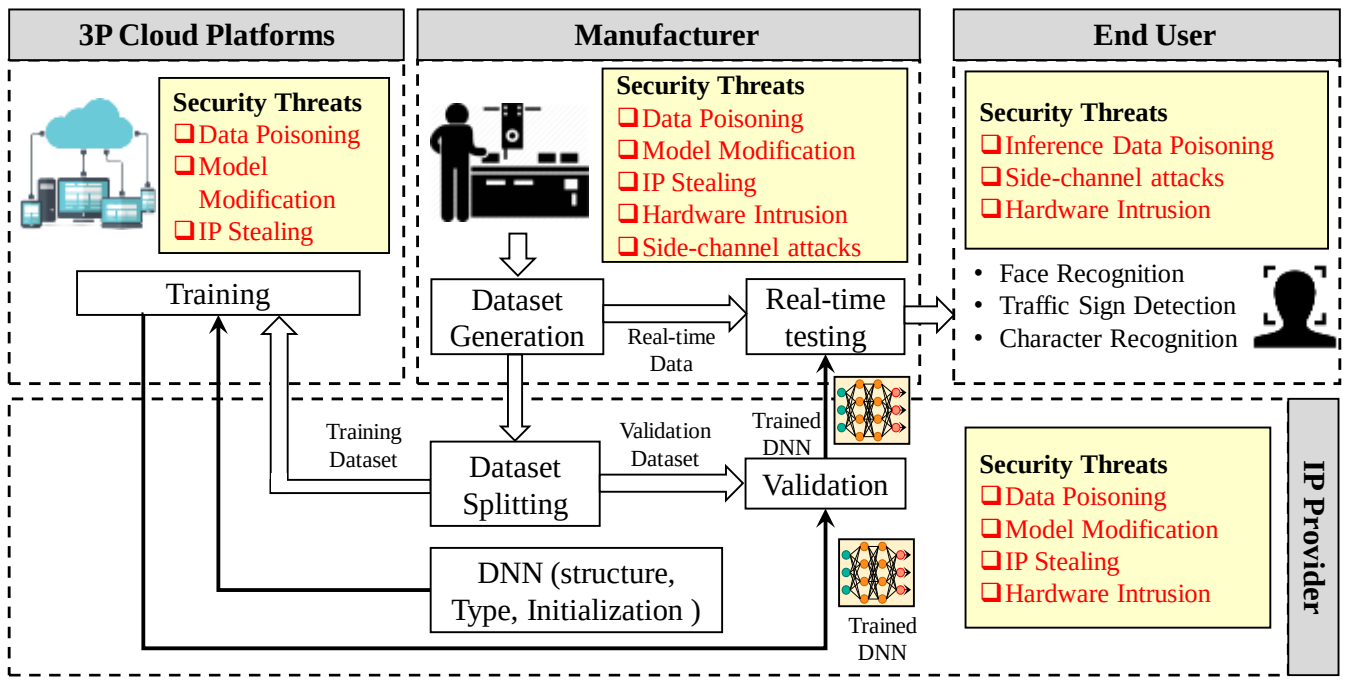

**Fig. 3: Security threats with respect to different actors involved in the manufacturing cycle of ML-based Applications.**

Thus, based on the trustworthiness of all actors involved in the design/manufacturing cycle, there can be 15 possible threat models for ML-based systems. For instance, *if any of the above-mentioned factors are untrusted then it can be considered as a potential threat model.*

### B. *Security Threats*

Based on the above-mentioned threat models, each actor can exploit or introduce security vulnerabilities in the ML-based system. Therefore, for developing security measures, the next step after defining the possible threat model is to identify the potential threats, their activation methodologies (i.e., data manipulation, malicious hardware and software intrusions) and corresponding payloads (i.e., confidence reduction (ambiguity in classification), random or targeted misclassification). Therefore, with respect to the manufacturing cycle, the following possible security threats can be identified.

1) **Training:** During training, an attacker can poison the training dataset and can also manipulate the tools/architecture/model [15], e.g., adding parallel layers or neurons, to perform security attacks, as shown in Fig. 4. However, in case of outsourced training, remote side-channel or cyber-attacks can be used to steal the IPs.

2) **Hardware Implementation:** Similarly, manipulation of the hardware implementation of the trained ML model (hardware Trojans) and IP stealing (i.e., side-channel, remote cyber-attacks) can be performed at hardware level [15], as shown in Fig. 4.

3) **Inference:** In this phase, there can be the following types of attacks:

   a) The user can attack (i.e., side-channel, remote cyber-attacks) the deployed ML-based system to steal the IP.

   b) Other possible attackers can be in-direct beneficiaries who can either manipulate the inference data or intrude the hardware [19]. Moreover, attackers can also perform side-channel attacks for IP stealing (see Fig. 4).

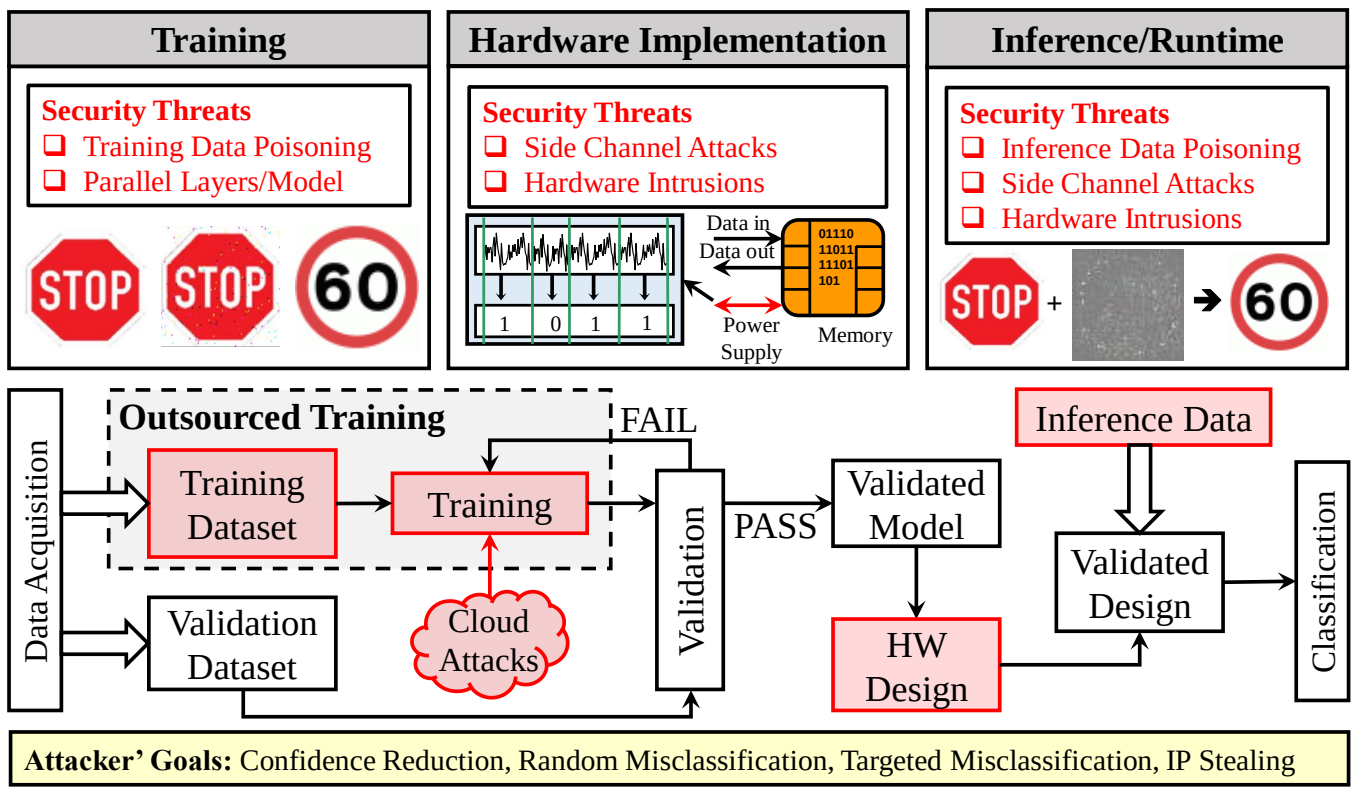

**Fig. 4: An Overview of Security Threats/Attacks and their respective payloads for Machine Learning Algorithms during Training, Inference, and their respective Hardware Implementations [15].**

## III. Security Analysis of ML-Based Systems

To illustrate the significance of security vulnerabilities in ML-based systems, we present a detailed analysis of some of the most common security vulnerabilities in ML, i.e., *data poisoning* (during training) and *adversarial examples* (during inference).

### A. *Training Attacks*

Training is one of the fundamental steps in developing ML-based systems and requires a lot of computational resources that encourages outsourced training on 3P cloud platforms. However, outsourcing comes with security vulnerabilities, i.e., data poisoning, IP (training data or trained model) stealing attacks or intrusions in baseline ML models/algorithms.

#### 1) *Random Misclassification Attack*

To demonstrate the security vulnerabilities during the training, we demonstrate a random misclassification attack through data poisoning in the MNIST [20] dataset during the training of LeNet [21] and in the German Traffic Sign Recognition Benchmarks (GTSRB) dataset [23] during the training of VGGNet [22], respectively. However, to perform the data poisoning attack on MNIST and GTSRB, the following research challenges need to be addressed:

1) What is the *optimal intensity of data poisoning (noise)* to perform random misclassification while maintaining the testing/targeted accuracy?
2) What is the optimal *number of intruded samples*?

To address the above-mentioned challenges, we analyzed LeNet for MNIST dataset with different number of intruded samples (1% to 40%) and with different noise types and intensities (i.e., *salt & pepper* and *Gaussian noise*), as shown in Fig. 5. The analysis shows that even with 1% (700/70000) of intruded samples and minimum salt & pepper noise (i.e., 10), the Top1 error of the LeNet is 0.8% which is not acceptable in testing the LeNet.

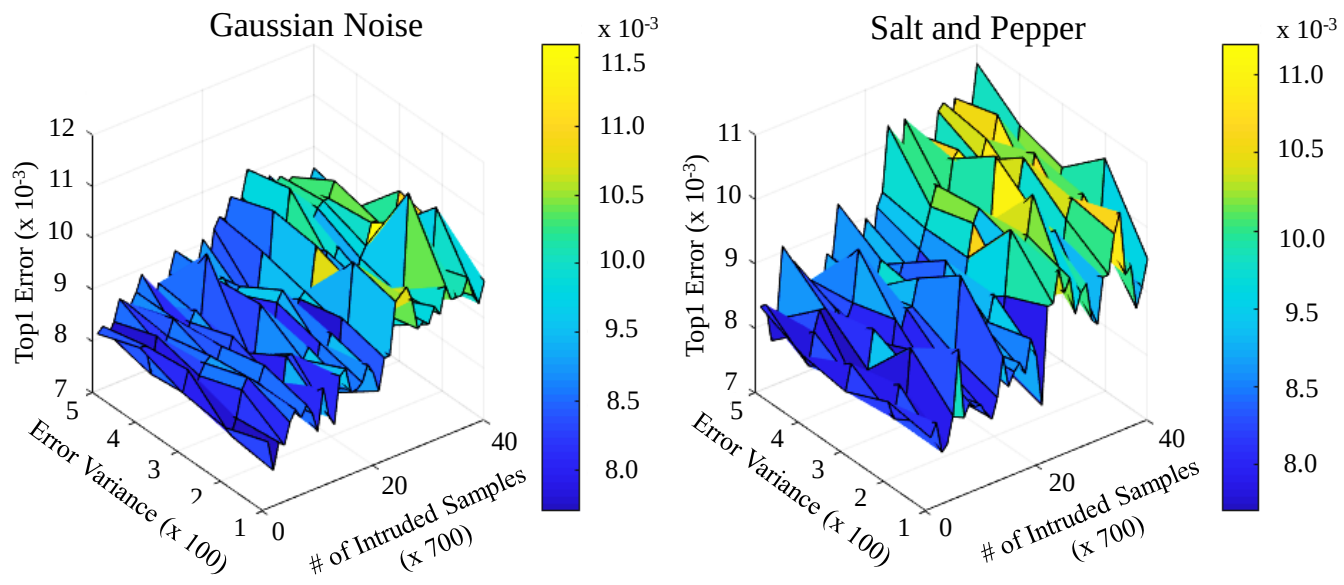

**Fig. 5: Impact of noise intensity (Salt & Pepper and Gaussian) on LeNet for MNIST dataset with 1% to 40% intruded number of samples.**

Therefore, to incorporate the effect on inference accuracy, we propose an attack which does not intrude the dataset, but it extends the dataset with certain malicious samples, as shown inFig. 6. To illustrate the effectiveness of this attack, we implement this attack on LeNet and VGGNet for MNIST and GTSRB datasets, respectively, and compare it with traditional dataset poisoning attacks.

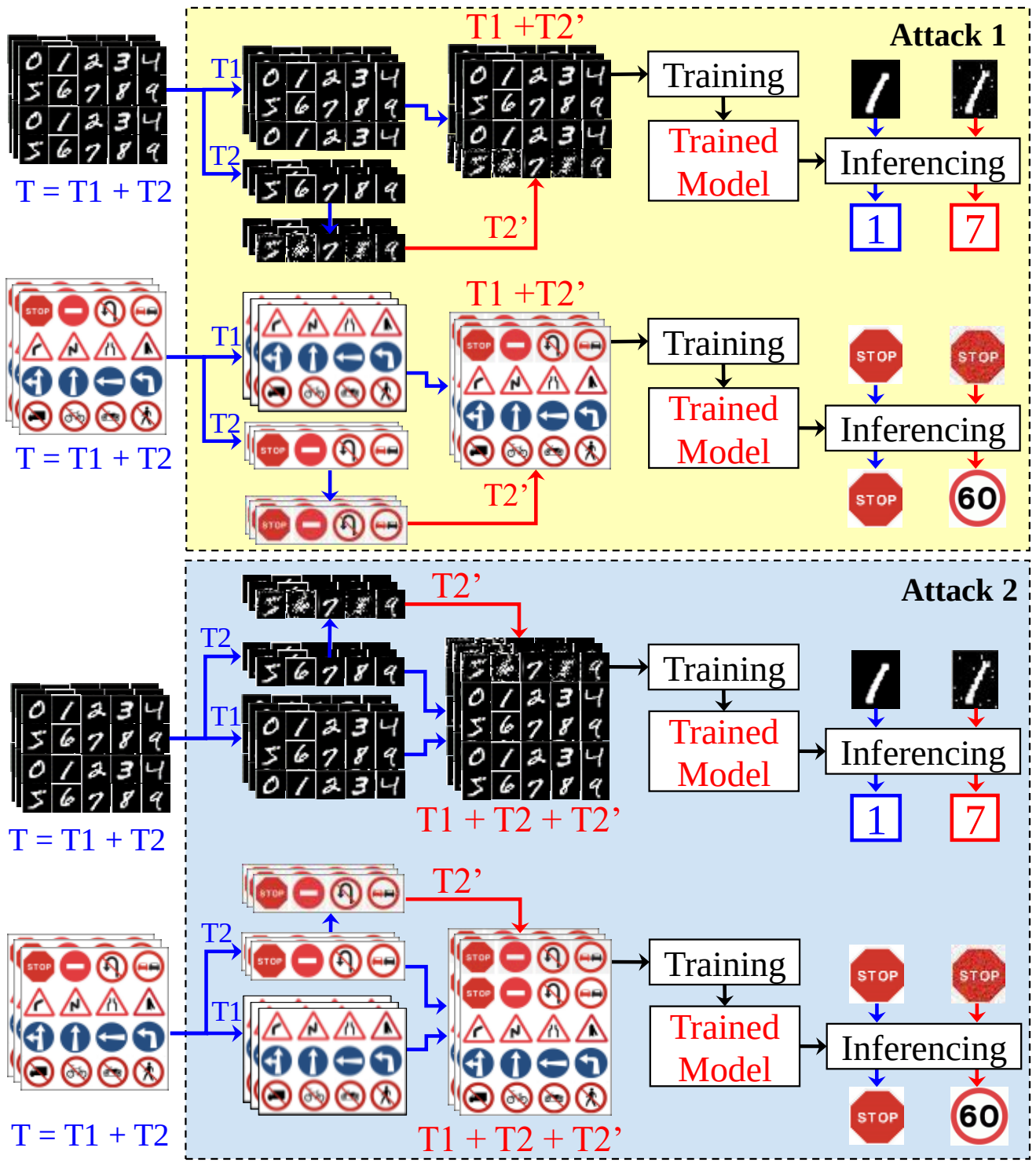

**Fig. 6: Our experimental setup for random misclassification attacks on LeNet and VGGNet for MNIST and GSRB dataset.**

**LeNet with MNIST:** First, we demonstrate the proposed attack (Attack 2 in Fig. 6) by training LeNet with only 3% (2100) intruded samples that are appended with MNIST. Similarly, we also performed the traditional attack (Attack 1 in Fig. 6) in which LeNet was trained on MNIST with similar number of intruded samples (3%, i.e., 2100). To demonstrate the effectiveness of these attacks, we perform the inference on intruded and un-intruded LeNets, as shown in Fig. 7. The analysis of these attacks shows that in both attacks the intruded LeNet randomly misclassifies the poisoned data sample (with label 0 as 8). However, the effect on inference accuracy of the proposed attack (1.3%) is less as compared to the traditional data poisoning attack (1.8%), as shown in Fig. 7.

**VGGNet with GTSRB:** Similarly, we also performed the proposed training data poisoning attack (Attack 2 in Fig. 6) and traditional attack (Attack 1 in Fig. 6) on VGGNet with GTSRB, as shown in Fig. 10. The experimental analysis shows that the impact of the proposed attack on inference accuracy is 20% less than the traditional attacks, as shown by the output distribution of 4% Salt & pepper noise in Fig. 10.

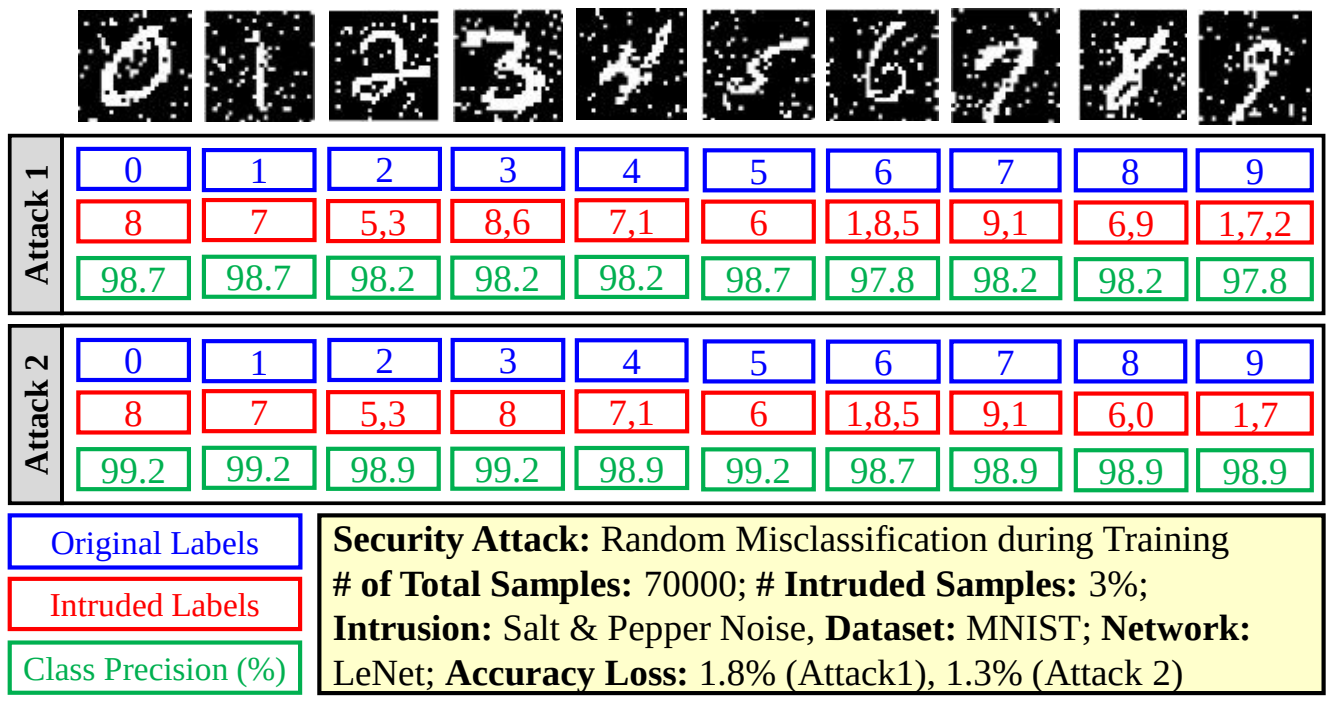

**Fig. 7: Random misclassification attack (i.e., introducing Salt & Pepper noise in 3% data samples of the MNIST dataset) on LeNet during the training phase.**

Based on the experimental analysis, we can conclude that randomly introduced noise in the training samples can be destructive because it reduces the confidence or misclassifies the input data. *For example, the collision avoidance in autonomous vehicles can be fooled by performing the random misclassification which leads to accidents.*

## B. *Inference Attacks*

In the development/manufacturing cycle of ML-based systems, like the traditional systems, the inference stages of ML algorithms come with security vulnerabilities, i.e., manipulation of data acquisition block, communication channels and side-channel analysis to manipulate the inference data and leaking IP (inference data and trained model). Remote cyber-attacks and side-channel attacks come with high computational costs and are therefore less frequently used. Consequently, to illustrate the impact of security vulnerabilities on inference, we demonstrate a couple of inference data poisoning attacks, i.e., adversarial examples (Limited-memory Broyden–Fletcher–Goldfarb–Shanno (L-BFGS) Method [24][25] and Fast Gradient Sign Method (FSGM) [26][27]). In this experimental analysis, we consider a threat model in which an attacker has the access to the dataset (images) right after the camera, as shown in Fig. 8.

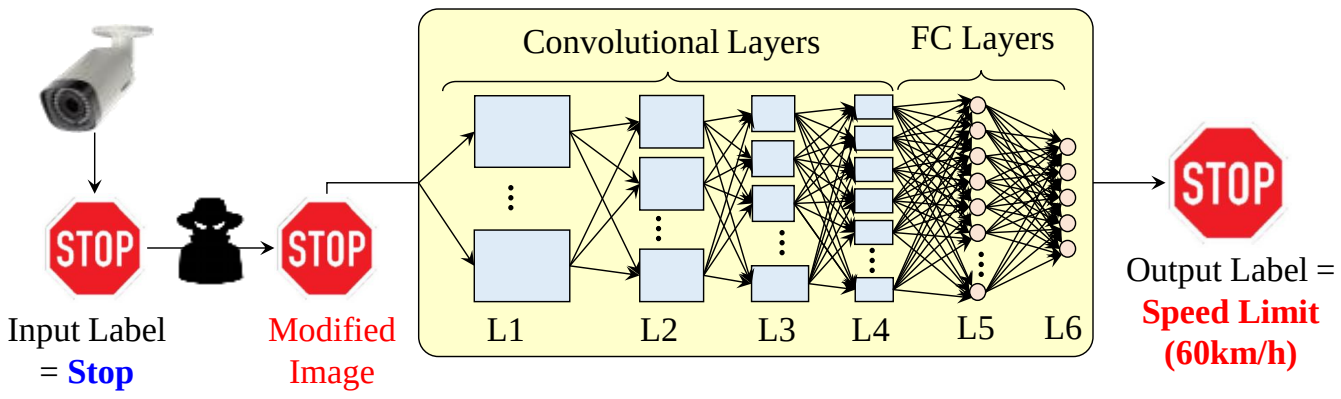

**Fig. 8: Threat model and experimental setup for inference attacks, i.e., adversarial examples.**

### 1) *Adversarial Examples*

Based on the threat model (Fig. 8), there can be several possible attacks. However, one of the most common attacks is to generate the adversarial examples [27][28]. The key goal of any adversarial example is to add an imperceptible noise into the data (images) that can force the ML algorithm to misclassification. To achieve this goal, an adversarial example typically follows the two-step methodology, as discussed below and shown in Fig. 9.

1) In the first step, an attacker chooses the target image/images or target output class/classes (in case of targeted misclassification) and defines the optimization goals, i.e., correlation coefficients, accuracy or other parameters to analyze imperceptibility.
2) In the second step, a random noise is introduced in the target image to compute the imperceptibility based on the defined optimization goals. If optimal imperceptibility is achieved, then the intruded image is considered as an adversarial image; otherwise, the noise is updated based on imperceptibility parameters and a new image is generated.

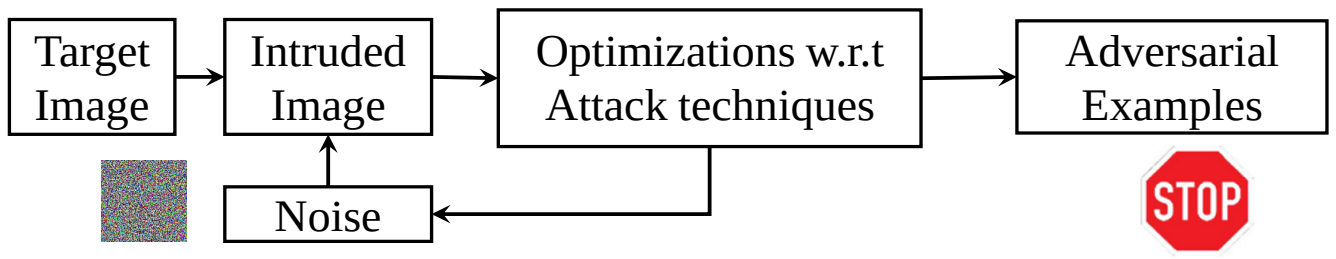

**Fig. 9: Basic methodology to generate an adversarial example.**

To analyze the impact of adversarial examples on inference, in this paper, we demonstrate two of the most commonly used adversarial attacks from the open-source *Cleverhans* library [29][30], i.e., L-BFGS Method [24][25] and Fast Gradient Sign Method (FSGM) [26][27].

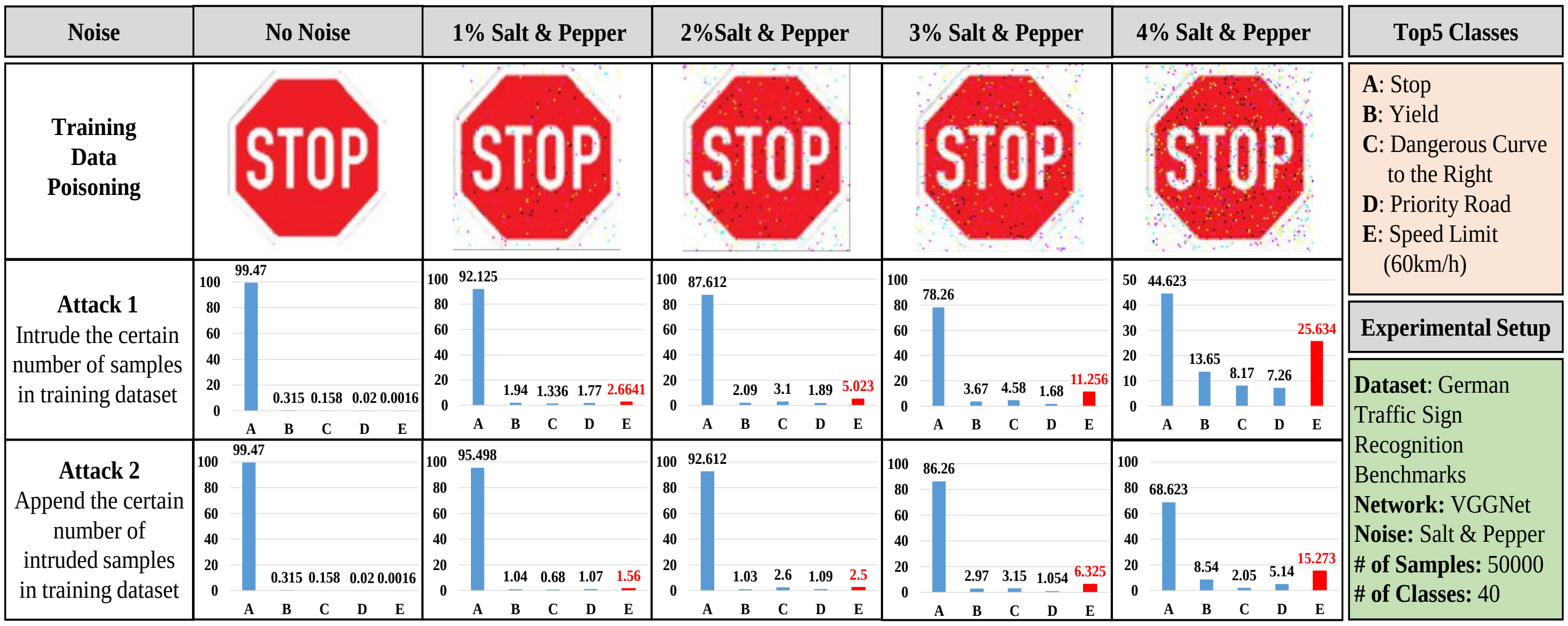

**Fig. 10: Random misclassification attack, i.e., introducing the salt and pepper noise in German Traffic Sign Recognition Benchmarks (GTSR).**

basic principle of the L-BFGS method is to achieve the optimization goal as defined in Equation 1.

$$Minimize \ \|noise\|_2 \ \rightarrow \ f(x + noise) \ \neq \ f(x) \quad (1)$$

Where, noise represents the perturbations and minimizing it represents its imperception. To illustrate the effectiveness of this method, we demonstrated this attack on the VGGNet trained on the GTSRB, as shown in Fig. 11. This experimental analysis shows that by introducing adversarial noise to the image, the input is misclassified, i.e., from a stop sign to speed limit 60km/h.

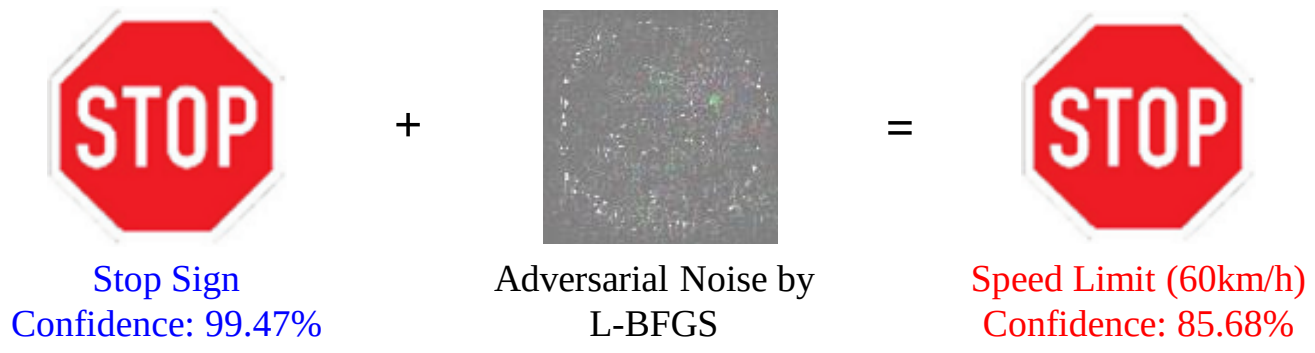

**Fig. 11: An adversarial image generated by L-BFGS Method.**

**Fast Gradient Sign Method (FSGM):** Although the L-BFGS method generates adversarial examples with imperceptible noise, it utilizes a basic linear search algorithm to update the noise for optimization which makes it computationally expensive and slow [26]. Therefore, Goodfellow et al. proposed a Fast Gradient Sign Method to generate adversarial examples which is faster and requires less computations as it performs one step gradient update along the direction of the sign of gradient at each pixel [27]. Their proposed imperceptive noise can be defined as:

$$\eta \ = \ \epsilon \, \nabla_x \, J(\theta, x, f) \qquad (2)$$

Where, $\epsilon$ and $\eta$ are the magnitude of the perturbation and the imperceptible noise, respectively. $J$ is the cost minimizing function (based on original image $x$, classification function $f$ and cost with respect to target class $\theta$) obtained through stochastic gradient descent. The generated adversarial example can be computed by adding $\eta$ into the targeted image. To analyze this vulnerability, we demonstrated this attack on VGGNet trained on GTSRB, as shown in Fig. 12.

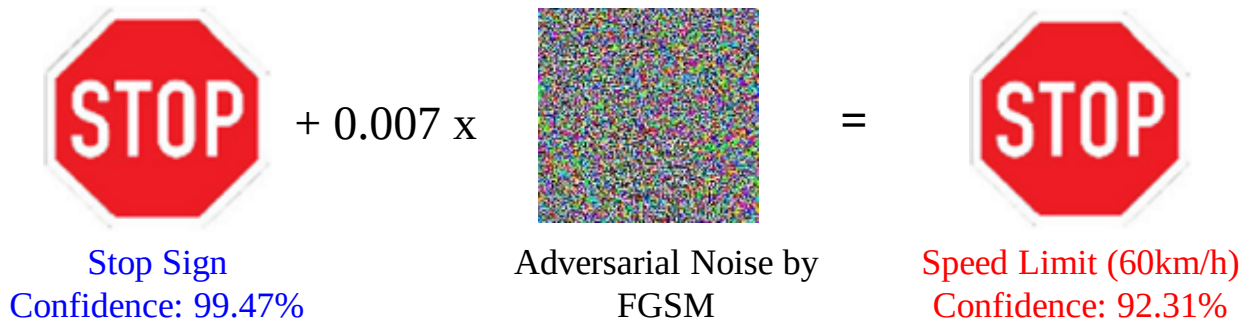

**Fig. 12: An adversarial image generated by Fast Gradient Sign Method. In this experiment, we assume the value for "$\epsilon$" is 0.007 to make it imperceptible, as mentioned in [27].**

## IV. Research Challenges for Design and Development of ML-Based System/Applications

Based on the above-mentioned security threats/vulnerabilities (Section II.B), the following research challenges need to be addressed for developing secure/robust ML-based systems:

1) How to *securely generate the training dataset* and ensure its privacy (especially, in outsourced design/implementation)?
2) How to *obfuscate the training dataset and hyper-parameters* of the underlying ML algorithm to ensure privacy during the outsourced training period?
3) How to ensure the *security of the data acquisition* during the inference stage?
4) How to *ensure the security of the pre-processing, ML algorithm and post processing hardware implementation*?
5) *How to validate the correctness and fairness* of the ML hardware implementation of the trained ML model?
6) *How to protect and obfuscate the IPs (i.e., trained ML model, dataset)* from IP stealing attacks, i.e., side channel, remote cyber-attacks, shared cache attacks?
7) *How to securely execute ML algorithms* on third-party hardware accelerators?

## V. Countermeasures for Security Vulnerabilities in Machine Learning

To address the above-mentioned challenges (Section IV), several countermeasures have been proposed, i.e., interactive proof (SafetyNets [33][34]), privacy-preserving predictions (SecureML [31], DeepNano [32]), encryption of dataset and trained ML models (CryptoNets [36]), classifier protections etc. In this section, we briefly discuss the several possible countermeasures.

Table 1 shows the summary of the potential security vulnerabilities and their corresponding countermeasures. Based on the manufacturing cycles, these countermeasures can be classified into the following three categories:

**Table 1: Summary of the potential security attacks and respective possible countermeasures with respect to manufacturing/design cycle.**

| | Security Vulnerabilities | Potential Countermeasures |
|---|---|---|
| **Training** | *Training Data Manipulation* | • Data Encryption,<br>• Redundant Outsourced Training<br>• Transfer Learning (local training of trained model which is obtained by outsourced training) |
| | *Cloud IP Stealing Attacks* | • Data Encryption<br>• Baseline ML-Model Obfuscation<br>• HW/SW Side-channel Analysis<br>• Online communication monitors<br>• Cyber Security Measures |
| **HW Implementation** | *Hardware Trojans* | • Online Monitoring<br>• HW/SW Side-channel Analysis<br>• Formal Verification and validation of hardware implementation<br>• Online property checkers |
| | *SC- IP Stealing* | • HW/SW Side-channel Analysis<br>• SC parameter-based runtime monitoring setup<br>• Online property checkers |
| | *Cyber-Attacks for IP Stealing* | • Online communication monitors<br>• Cyber Security Measures |
| **Inference** | *Inference Data Manipulation* | • Data Encryption<br>• Sophisticated Pre-processing |
| | *HW/SW IP stealing Attacks* | • Online communication monitors<br>• Cyber Security Measures<br>• HW/SW Side-channel Analysis<br>• SC parameter-based runtime monitoring setup |

1) **Training:** Depending upon the targeted vulnerabilities and corresponding methodologies, there are several possible countermeasures:
   a) *Encryption*: In this approach, the training dataset set is encrypted before training (local/outsource), i.e., cryptoNets [36][37][38]. However, this countermeasure comes with additional hardware for encrypting the inference dataset.
   b) *Transfer Learning-based Local Training*: To mitigate the manipulations of weight/model, the dataset is split into two parts, one for outsourced training and other one is to locally train the model, which can be used to overwrite the outsourced trained weights/model by performing the transfer learning [39][40].
   c) *Redundant Training*: To mitigate the manipulation of trained model/data, the training is outsourced to multiple 3P cloud

platforms. Triple/multi-modular redundancy is used to identify the intrusions while testing [41].

2) **Hardware Implementation:** Hardware intrusions in ML-based systems are similar to traditional hardware-attacks, therefore, typical hardware security techniques can be applied, i.e., runtime anomaly detection using side-channel and communication analysis, formal method-based analysis ([42]-[44]), and traditional obfuscation techniques to mitigate hardware IP stealing.

*3)* **Inference:** Similarly, traditional obfuscation techniques, runtime anomaly detection, side channel analysis and security measures for remote cyber-attacks [45] can be applied to mitigate the IP stealing. data, trained model and hardware manipulation. Development and selection of appropriate security measures for inference is a very complex and tedious process because of the system's energy and design constraints, especially in battery operated components of a CPS [46]. For example, to avoid the data poisoning attacks, encryption is one of the possible countermeasures, but due to limited energy resources applicability of this is limited.

## VI. Conclusion

In this paper, we first discuss and identify the possible security threats in machine learning with respect to threat models, attack methodologies and payloads. Moreover, to analyze the security vulnerabilities for identifying the potential countermeasures, we demonstrate some of the security threats (Training data poisoning and adversrial examples (L-BFGS and FSGM)) on the LeNet and the VGGNet for the MNIST and the German Traffic Sign Recognition Benchmarks (GTSRB), respectively. We also propose a training data poisoning attack which has relatively less impact on inference accuracy. Finally, we provide an overview of possible security measures and highlight respective research challenges in developing these security measures.